\pdfoutput=1

\documentclass[11pt]{article}

\usepackage[preprint]{coling}

\usepackage{times}
\usepackage{latexsym}

\usepackage[T1]{fontenc}

\usepackage[utf8]{inputenc}
\usepackage[main=english,ukrainian]{babel}

\usepackage{microtype}

\usepackage{inconsolata}

\usepackage{graphicx}

\usepackage{booktabs}
\usepackage{color, colortbl}
\usepackage{enumitem}
\usepackage{caption}
\usepackage{adjustbox}
\usepackage{amssymb}
\usepackage{utfsym}
\usepackage{mathtools}
\usepackage{amsmath}
\usepackage{subcaption}

\usepackage{xcolor}

\definecolor{Gray}{gray}{0.9}

\title{Cross-lingual Text Classification Transfer: \\ The Case of Ukrainian}

\author{Daryna Dementieva, Valeriia Khylenko \and {\bf Georg Groh} \\
Technical University of Munich, School of Computation, Information and Technology, Germany\\ 
\href{mailto:daryna.dementieva@tum.de}{\texttt{\small daryna.dementieva@tum.de}},
\href{mailto:l.khylenko@gmail.com}{\texttt{\small l.khylenko@gmail.com}},
\href{mailto:grohg@in.tum.de}{\texttt{\small grohg@in.tum.de}}
}

\begin{document}
\maketitle
\begin{abstract}
Despite the extensive amount of labeled datasets in the NLP text classification field, the persistent imbalance in data availability across various languages remains evident.
To support further fair development of NLP models, exploring the possibilities of effective knowledge transfer to new languages is crucial. 
Ukrainian, in particular, stands as a language that still can benefit from the continued refinement of cross-lingual methodologies. Due to our knowledge, there is a tremendous lack of Ukrainian corpora for typical text classification tasks, i.e., different types of style, or harmful speech, or texts relationships. However, the amount of resources required for such corpora collection from scratch is understandable. In this work, we leverage the state-of-the-art advances in NLP, exploring cross-lingual knowledge transfer methods avoiding manual data curation: large multilingual encoders and translation systems, LLMs, and language adapters. We test the approaches on three text classification tasks---toxicity classification, formality classification, and natural language inference (NLI)---providing the ``recipe'' for the optimal setups for each task. \\ \textcolor{red}{\textit{Warning: This paper contains offensive texts that only serve as illustrative examples.}}
\end{abstract}

\section{Introduction}

\begin{table}[ht!]
    \centering
    \scriptsize
    \begin{tabular}{p{1.70cm}|p{5.3cm}}
    \toprule

        \multicolumn{2}{c}{\textbf{Toxicity Classification}} \\
        \midrule 
        Toxic \newline \newline \newline Non-toxic & \foreignlanguage{ukrainian}{Послухайте вас, п*дики розблоковують мене або я вас усіх вб'ю.} \newline \textcolor{gray}{\scriptsize{\textit{listen u wikipedia f*gs unblock me or i kill u all}}} \newline\foreignlanguage{ukrainian}{Справедливе обурення завжди смішне.} \newline \textcolor{gray}{\scriptsize{\textit{righteous indignation always funny}}} \\
        \midrule 

        \multicolumn{2}{c}{\textbf{Formality Classification}} \\
        \midrule 
        Formal \newline \newline \newline \newline Informal & \foreignlanguage{ukrainian}{Іноді, якщо добро переважає зло, то труднощі того варті.} \newline \textcolor{gray}{\scriptsize{\textit{Sometimes, if the good outweighs the bad, then the difficulties are worth it.}}} \newline \foreignlanguage{ukrainian}{Я знаю, що ви бачили смішніше, але це все ж робить мене безглуздим.} \newline \textcolor{gray}{\scriptsize{\textit{I know i know u seen funnier but it still makes me laff :)}}} \\
        \midrule 

        \multicolumn{2}{c}{\textbf{Natural Language Inference (NLI)}} \\
        \midrule 
        Premise \newline \newline \newline Hypothesis \newline \newline Label & \foreignlanguage{ukrainian}{Три пожежники виходять з станції метро.} \newline \textcolor{gray}{\scriptsize{\textit{Three firefighter come out of subway station.}}} \newline \foreignlanguage{ukrainian}{Три пожежники грають в карти в пожежному відділенні.} \newline \textcolor{gray}{\scriptsize{\textit{Three firefighters playing cards inside a fire station.}}} \newline contradiction\\

        \bottomrule
    \end{tabular}
    \caption{Samples of the considered tasks.}
    \label{tab:intro_examples}
\end{table}

In recent years, the NLP community has shifted its focus beyond monolingual English models, placing greater emphasis on developing fair and equitable multilingual NLP technologies. Even if the state-of-the-art language models like, for instance, BERT~\cite{devlin-etal-2019-bert}, RoBERTa~\cite{liu2019roberta}, T5~\cite{raffel2020exploring}, BART~\cite{lewis2019bart} were firstly pre-trained only for English, then their multilingual versions also appeared---mBERT, XLM-RoBERTa~\cite{conneau2019unsupervised}, mT5~\cite{xue2020mt5}, and mBART~\cite{tang2020multilingual}. Also, the next generation of multilingual models family like BLOOMz~\cite{muennighoff2022crosslingual} were introduced. The translation systems also received a boost with the recent NLLB model covering 200 languages~\cite{costa2022no}. Finally, Large Language Models (LLMs) pre-trained on huge corpora opened the possibilities of emerging abilities~\cite{wei2022emergent} not only for new tasks, but even for languages.

\begin{figure*}[th!]
    \centering
    \includegraphics[width=\textwidth]{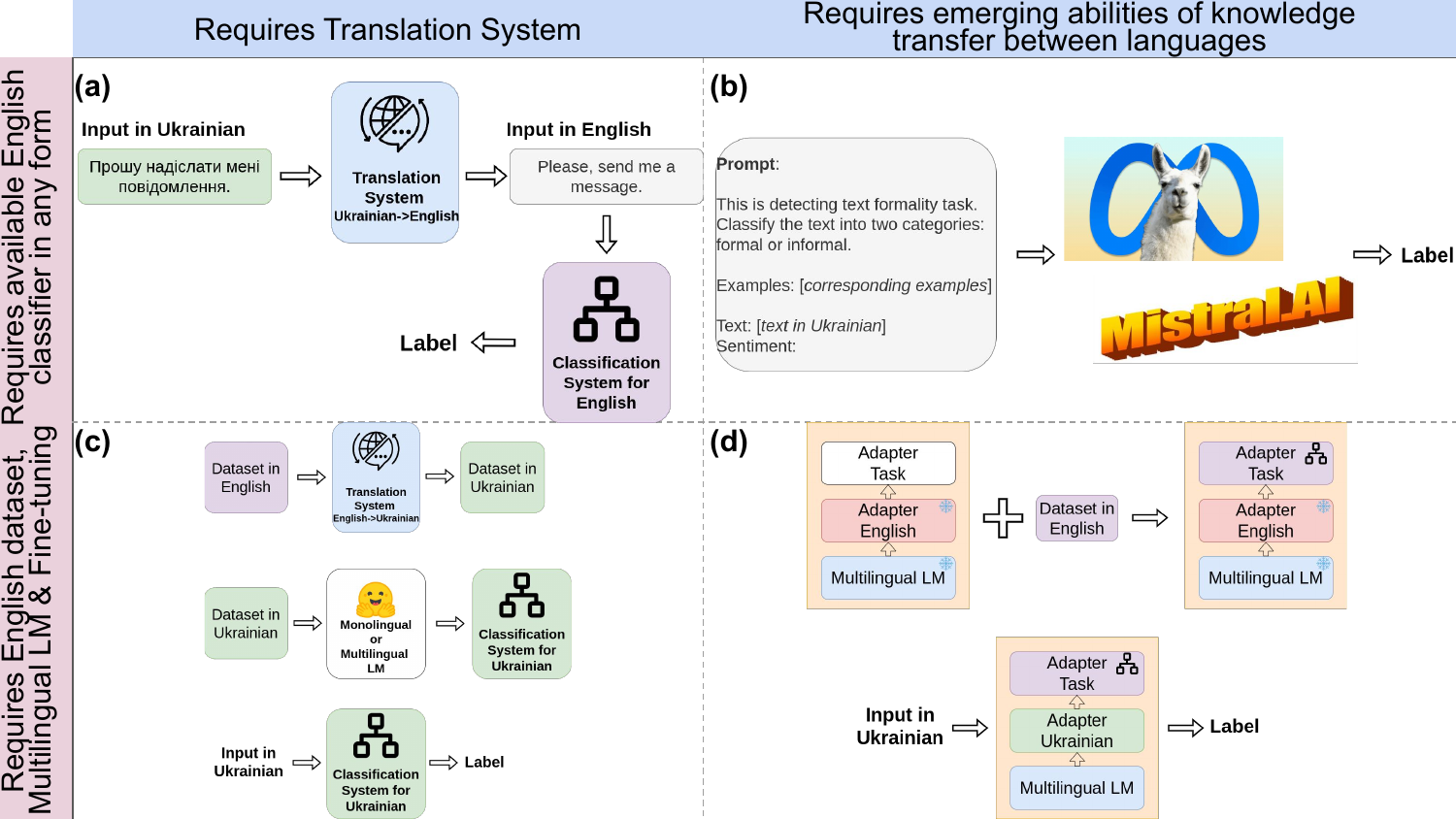}
    \caption{\textbf{Cross-lingual knowledge transfer approaches} explored for Ukrainian texts classification: (a) Backtranslation; (b) LLM Prompting; (c) Training Corpus Translation; (d) Adapter Training. The approaches requires different resources availability and dependence on a translation system.}
    \label{fig:all_methods}
\end{figure*}

Nevertheless, the scope of language coverage remains constrained. Furthermore, as of our current understanding, there exists a gap in the formal exploration of the effectiveness of before mentioned multilingual LMs for obtaining an NLP text classification system for a new language. With this work, we are aiming to close this gap exploring cross-lingual knowledge transfer approaches for the Ukrainian language. Thus, our contributions are the following:
\begin{itemize}
    \item We are the first to explore four types of cross-lingual text classification transfer approaches---Backtranslation, LLMs Prompting, Training Corpus Translation, and Adapter Training--applying them for Ukrainian;
    \item As a result, we design the first of its kind datasets and models for Ukrainian texts classification for three tasks---toxicity classification, formality classification, and NLI (Table~\ref{tab:intro_examples});
    \item The results are obtained on both synthetic translated and semi-natural test sets providing insights into the methods effectiveness.
\end{itemize}
All the obtained data and models are available for the public usage online.~\footnote{\href{https://huggingface.co/ukr-detect}{https://huggingface.co/ukr-detect}}


\section{Related Work}

The usual case for cross-lingual transfer setup is when data for a specific task is available for English but none for the target language.
One of the approaches used to address the lack of training data is the translation approach. Such approached was already explored for the sentiment analysis~\cite{DBLP:journals/apin/KumarPR23} and offensive texts classification \cite{el2022multilingual,DBLP:journals/csse/WadudMSNS23}. In the related domain, Ukrainian bullying detection system was developed based on the translated English data in \cite{oliinyk2023low}.

In \cite{dong-de-melo-2019-robust}, robust self-learning framework was designed based on the incorporation of unlabeled non-English samples during the fine-tuning phase of pretrained multilingual representation models. 
To decrease the size of the trained models parameters, Adapter layers were introduced in \cite{DBLP:conf/icml/HoulsbyGJMLGAG19} as a more efficient way of downstream tasks models fine-tuning and language adjustment. It was successfully tested for token-level classification transfer in~\cite{rathore-etal-2023-zgul} for several Asian and European languages. Finally, zero-shot and few-shot prompting of LLMs~\cite{winata-etal-2022-cross} can be a promising approach to obtain a baseline classifier for a language. However, none of the work yet explored the main cross-lingual transfer approaches for Ukrainian.

Although training data for various classification tasks in Ukrainian remains limited, the community has made substantial progress in token-level understanding tasks, machine translation, and increasing the presence of Ukrainian in pre-trained datasets. Thus, UberText 2.0~\cite{chaplynskyi-2023-introducing} covers NER detection tasks, legal texts in Ukrainian, and other massive data from various resources--news, Wikipedia, fiction. Another source for Ukrainian data is the parallel OPUS corpus~\cite{tiedemann-2012-parallel}. Moreover, the Spivavtor dataset~\cite{saini-etal-2024-spivavtor} has been introduced to support the instruction-tuning of Ukrainian-specialized LLMs.


\begin{table}[ht!]
\centering
\scriptsize
\begin{tabular}{@{}p{1cm}|lll@{}}
\toprule
                                  & Toxicity dataset                                                                          & Formality dataset                                                                         & NLI dataset                                                                                                          \\ \midrule
Train        & \begin{tabular}[c]{@{}l@{}}total: 24616\\ toxic: 12307\\ non-toxic: 12309\end{tabular} & \begin{tabular}[c]{@{}l@{}}total: 209124\\ formal: 104562\\ informal: 104562\end{tabular} & \begin{tabular}[c]{@{}l@{}}total: 549361\\ neutral: 182762\\ contradiction: 183185\\ entailment: 183414\end{tabular} \\ \midrule
Val          & \begin{tabular}[c]{@{}l@{}}total: 4000\\ toxic: 2000\\ non-toxic: 2000\end{tabular}    & \begin{tabular}[c]{@{}l@{}}total: 10272\\ formal: 4605\\ informal: 5667\end{tabular}      & \begin{tabular}[c]{@{}l@{}}total: 9842\\ neutral: 3235\\ contradiction: 3278\\ entailment: 3329\end{tabular}         \\ \midrule
Test       & \begin{tabular}[c]{@{}l@{}}total: 52294\\ toxic: 5800\\ non-toxic: 46494\end{tabular}  & \begin{tabular}[c]{@{}l@{}}total: 4853\\ formal: 2103\\ informal: 2750\end{tabular}       & \begin{tabular}[c]{@{}l@{}}total: 9824\\ neutral: 3219\\ contradiction: 3237\\ entailment: 3368\end{tabular}         \\ \midrule
\begin{tabular}[t]{@{}p{1cm}@{}}Semi-natural Test \end{tabular} & \begin{tabular}[t]{@{}l@{}}total: 4214\\ toxic: 2114\\ non-toxic: 2088\end{tabular}    & \begin{tabular}[t]{@{}l@{}}total: 3000\\ formal: 1500\\ informal: 1500\end{tabular}       & \begin{tabular}[t]{@{}l@{}}total: 901\\ neutral: 300\\ contradiction: 300\\ entailment: 301\end{tabular}             \\ \bottomrule
\end{tabular}
\caption{Statistics of the tasks datasets: train/val/test splits were obtained via translation from the corresponding English datasets; also, we constructed semi-natural test sets to evaluate the models under conditions resembling real-life scenarios.}
\label{tab:dataset_stat}
\end{table}

\section{Methodology}

We take into consideration four cross-lingual knowledge transfer methods (Figure~\ref{fig:all_methods}): (i)~Backtranslation; (ii)~LLM Prompting; (iii)~Training Corpus Translation; (iv)~Adapter Training. To consider the most popular scenario, everywhere, we assume English as a resource-rich available language.

\paragraph{Backtranslation} 
One of the natural baselines can be to translate the input in Ukrainian into English and then utilize such an English classifier for the task. Such a Backtranslation approach does not require fine-tuning; however, it depends on the constant calls of external models---a translation system and an English classifier.

\paragraph{LLM Prompting} The next approach that as well does not require fine-tuning is prompting of LLMs. Current advances in generative models showed the feasibility of transforming any NLP classification task into text generation task \cite{chung2022scaling,DBLP:conf/emnlp/AlySLZW23}. Thus, the prompt can be designed in a zero-shot or a few-shot manner requesting the model to answer with the label. While LLMs were already tested for a hate speech classification task for multiple languages~\cite{das2023evaluating}, there were no yet experiments for any text classification task for the Ukrainian language which might be underrepresented in such models. We provide the final designs of our prompts in Appendix~\ref{sec:app_llm_prompts}.

\paragraph{Training Corpus Translation} To avoid the permanent dependence on a translation system per each request, we can translate the whole English dataset and, as a result, get synthetic training data for the task. Then, a downstream task fine-tuning is possible. This approach's main advantage is that there are no external dependencies during the inference time, but it requires computational resources for fine-tuning. Moreover, some class information might vanish after translation and will not be adapted for the target language. 

\paragraph{Adapter Training} Finally, the most parameter-efficient approach involves employing language-specific Adapter layers~\cite{pfeiffer-etal-2020-adapterhub}. We followed the regular pipeline of cross-lingual transfer with language Adapters.\footnote{\href{https://github.com/adapter-hub/adapters/blob/main/notebooks/04_Cross_Lingual_Transfer.ipynb}{https://github.com/adapter-hub/adapters/\\blob/main/notebooks/04\_Cross\_Lingual\_Transfer.ipynb}} Such a language Adapter, firstly, for the source language---English---can be added upon multilingual encoder. Everything remains frozen while fine-tuning of the another Adapter for the downstream task. Then, English Adapter is replaced with the Ukrainian one and an inference step for the task in the target language can be performed.

\section{Experimental Setup}

\begingroup
\renewcommand{\arraystretch}{1.1}
\begin{table*}[ht!]
\footnotesize
\centering

\begin{tabular}{p{3.65cm}|c|c|c|c|c|c|c|c}
\toprule

& \textbf{Acc} & \textbf{Pr} & \textbf{Re} & \textbf{F1} & \textbf{Acc} & \textbf{Pr} & \textbf{Re} & \textbf{F1} \\ \hline

& \multicolumn{4}{c|}{{Translated Test Set}} & \multicolumn{4}{c}{{Semi-natural Test Set}} \\ \hline \hline

\multicolumn{9}{c}{\textbf{Toxicity Classification}} \\ \hline

Mistral Prompting &   0.86 &
  \underline{0.68} &
  0.74 &
  0.70 &
  \textbf{0.76} &
  \underline{\textbf{0.81}} &
  \textbf{0.76} &
  \textbf{0.75}  \\
Backtranslation & \multicolumn{4}{c|}{---}  & 0.63 & 0.76 & 0.56 & 0.58  \\
\hline
Adapter Training & \underline{\textbf{0.87}} & 0.66 & 0.63 & 0.65 & 0.58 & 0.66 & 0.58 & 0.52 \\
XLM-R-finetuned & 0.81 & \textbf{0.68} & \underline{\textbf{0.86}} & \underline{\textbf{0.70}}  & \underline{\textbf{0.77}} & \textbf{0.79} & \underline{\textbf{0.77}} & \underline{\textbf{0.77}} \\

\hline \hline

\multicolumn{9}{c}{\textbf{Formality Classification}} \\ \hline
%
Mistral Prompting &     \underline{0.64} &
  \underline{0.63} &
  \underline{0.64} &
  \underline{0.63} &
  \underline{\textbf{0.94}} &
  \underline{\textbf{0.94}} &
  \underline{\textbf{0.94}} &
  \underline{\textbf{0.94}}  \\
Backtranslation & \multicolumn{4}{c|}{---}  & 0.56 & 0.61 & 0.39 & 0.50  \\
\hline
 Adapter Training & \textbf{0.64} & \textbf{0.63} & 0.63 & \textbf{0.63}  & \textbf{0.71} & \textbf{0.71} & \textbf{0.71} & \textbf{0.71} \\
XLM-R-finetuned & 0.57 & 0.28 & 0.50 & 0.36  & 0.50 & 0.25 & 0.50 & 0.33 \\

\hline \hline

\multicolumn{9}{c}{\textbf{Natural Language Inference}} \\ \hline

Mistral Prompting &     0.56 &
  0.61 &
 0.56 &
  0.56 &
  \underline{\textbf{0.71}} &
  \underline{\textbf{0.72}} &
  \underline{\textbf{0.69}} &
  \underline{\textbf{0.69}}  \\
Backtranslation & \multicolumn{4}{c|}{---} & 0.40 & 0.41 & 0.63 & 0.33  \\
\hline
Adapter Training & 0.44 & 0.46 & 0.43 & 0.41  & 0.40 & 0.36 & 0.40 & 0.32 \\
XLM-R-finetuned & \underline{\textbf{0.82}} & \underline{\textbf{0.82}} & \underline{\textbf{0.82}} & \underline{\textbf{0.82}} & \textbf{0.48} & \textbf{0.46} & \textbf{0.46} & \textbf{0.42} \\

\bottomrule
\end{tabular}
\caption{Ukrainian Texts Classification results. We divide methods into two groups -- not requiring and requiring fine-tuning. Then, \textbf{bold} numbers denote the best results within the methods group and a test set, \underline{underline} -- overall best scores for the task.}
\label{tab:all_results}
\end{table*}
\endgroup

\paragraph{Tasks English Datasets} To test the approaches, we considered three text classification task and corresponding English datasets (Table~\ref{tab:dataset_stat}): (i) toxicity classification based on Jigsaw data~\cite{jigsaw} (we collapsed all labels except from ``non-toxic'' into one ``toxic'' class); (ii) formality classification with GYAFC~\cite{rao-tetreault-2018-dear}; (iii) NLI task on the benchmark dataset SNLI~\cite{snliemnlp2015}. We saved the original set splits. Translated data examples can be found in Appendix~\ref{sec:app_translated_data}.

\paragraph{Translation Systems Choice} To choose the most appropriate translation system, we took into consideration two opensource models---NLLB~\cite{costa2022no} and Opus~\cite{tiedemann-2012-parallel}. We randomly selected 50 samples per each dataset and asked 3 annotators (native speakers in Ukrainian) to verify the quality. For the annotators answers aggregation, we used the majority voting. As a result, we chose Opus translation system\footnote{\href{https://huggingface.co/Helsinki-NLP/opus-mt-en-uk}{https://huggingface.co/Helsinki-NLP/opus-mt-en-uk}} for toxicity classification as it preserves better the toxic lexicon, for others---NLLB.\footnote{\href{https://huggingface.co/facebook/nllb-200-distilled-600M}{https://huggingface.co/facebook/nllb-200-distilled-600M}} For the respected tasks. both systems achieved 90\% of qualitative translations based on the aggregated annotation results.

\paragraph{Ukrainian Texts Encoder Choice} For the Ukrainian texts encoder---for the Adapter training and the classifier fine-tuning---XLM-RoBERTa\footnote{\href{https://huggingface.co/FacebookAI/xlm-roberta-base}{https://huggingface.co/FacebookAI/xlm-roberta-base}}~\cite{conneau2019unsupervised} that was pre-trained including Ukrainian data has already been proven as a strong baseline for multiple languages~\cite{imanigooghari2023glot500}.

\paragraph{LLM Choice} For LLMs prompting, we experimented with couple setups (details in Appendix~\ref{sec:app_llms_exploration})
choosing Mistral\footnote{\href{https://huggingface.co/mistralai/Mistral-7B-v0.1}{https://huggingface.co/mistralai/Mistral-7B-v0.1}}~\cite{Jiang2023Mistral7}
as the most promising model (to this date) for Ukrainian texts processing.

\paragraph{Semi-natural Test Sets}
In addition to the translated test sets, we prepared tests sets based on automatic pre-processing of natural Ukrainian texts to assess the models in circumstances mirroring real-world scenarios. The texts examples are presented in Appendix~\ref{sec:app_natural_data}.

For toxicity, natural test part was collected from two sources: (i)~Ukrainian tweets corpus from~\cite{bobrovnik2019twt} where tweets were filtered based on toxic keywords from~\cite{tox_uk_lexicon}; (ii) additional non-toxic sentences were obtained from news and fiction UD Ukrainian IU dataset~\cite{udiu2016}.

Informal sentences in the formality natural test dataset were also from the tweets corpus, while formal sentences were collected from Ukrainian legal acts~\cite{legaltestuk2012} and EU acts in Ukrainian~\cite{euactsukr2012} corpora.

For the entailment label for NLI, also Ukrainian legal acts data was utilized, as well as open corpus of modern Ukrainian~\cite{chaplynskyi-2023-introducing}. Neutral sentences were taken from the fiction corpora~\cite{lanorgfiction2022}. Finally, contradiction pairs were constructed by the Ukrainian native speaker.

\section{Results}
The final results are presented in Table~\ref{tab:all_results}. We report primary text classification metrics: accuracy, precision, recall, and F1 scores. We report for Backtranslation results only on the semi-natural test sets (English SOTA comparison in Appendix~\ref{sec:app_english_sota}).

For toxicity classification, Mistral overcame Backtranslation within the baselines that do not require fine-tuning. However, the fine-tuned XLM-RoBERTA scores significantly superior on both test set types. Even if the training data were obtained from English that is less rich on morphological forms of toxic phrases, this model can be used as a strong toxicity detector baseline.

In contrast to the toxicity task, Adapter Training demonstrates the most reliable results for formality classification, whereas fine-tuning XLM-R was unsuccessful. This underperformance may be due to the loss of crucial information about formal and informal classes during the translation process. On the other hand, Mistral, which is primarily trained on English data, retained the necessary formality information and effectively transferred it to the target Ukrainian language.

For the NLI task, Mistral once again outperformed all baselines. However, there was a significant drop in XLM-RoBERTa's performance between the translated and natural test sets, likely due to domain differences, highlighting the need for native Ukrainian data in NLI tasks.

\section{Conclusion}
We presented the first-of-its-kind study of the cross-lingual transfer approaches for texts classification task tested on Ukrainian. Three tasks were considered---toxicity classification, formality classification, and natural language inference. We tested two zero-shot approaches---Backtranslation that depends on a translation system inference and LLM Prompting---and two approaches that require model fine-tuning---Adapter Training that updates only task-specific layer and Training Corpus Translation. As a result of our experiments, we obtained Ukrainian-translated datasets for the examined tasks, along with compiled semi-natural test sets for more realistic evaluations.

For the final recommendations, LLM prompting---particularly with Mistral---can serve as a solid baseline for Ukrainian texts processing, with the exception of toxicity classification. In that case, fine-tuned XLM-RoBERTa outperformed other approaches. However, aside from formality classification, the leading results for all tasks still show potential for improvement. Although we have introduced robust baselines for Ukrainian text classification, we strongly encourage further additional investigations using native Ukrainian data for these tasks.


\section*{Limitations}
In this work, we only explored three sentence level classification tasks. While the token-level classification for Ukrainian is already at a very good level~\cite{chaplynskyi-2023-introducing}, for sentence level there is still a room for improvement. We made a focus on the tasks which were already a field of expertise of the authors shading the light on the perspectives of modern methods utilization for Ukrainian. At the same time, there is a still a room for other texts classification tasks exploration.

Given resource constraints, our experiments only incorporated base and distilled versions of the models. Despite these limitations, the approaches we explored yielded promising results. However, employing models with more parameters could yield even stronger outcomes. Furthermore, for translation and LLMs prompting, we exclusively utilized open-source models. Exploring enterprise models could potentially offer the boost in the performance and more robust industrial solutions.

In conclusion, we opted to perform cross-lingual transfer from English, considering it as the most resource-rich language for the most general scenario. However, if resources such as datasets and models are accessible for languages closer to Ukrainian, such as Polish or Croatian, conducting cross-lingual transfer from these languages could potentially yield even better results.

\section*{Ethics Statement}
Although this study examines several cross-lingual classification methods with semi-natural test sets, it does not involve thoroughly exploring or properly annotating authentic Ukrainian data. Consequently, relying on translations or assumption-based data construction may introduce errors and noise in the data. Moreover, such datasets may not accurately reflect the current state of the Ukrainian language as used online. While our goal is to provide baselines and a foundation for further exploration, the proposed approaches must be carefully validated by stakeholders prior to real-world deployment.

The work primarily centers on Ukrainian language support, aiming to address its underrepresentation in the context of language development. We strongly believe, the obtained findings can server as an inspiration for promoting fairness in the development of other languages.
%
%
Overall, this work not only contributes to the advancement of Ukrainian language technology but also provides a blueprint for equitable language development practices that can be applied to other languages facing similar challenges.

\section*{Acknowledgments}

This article was supported by the Friedrich Schiedel Fellowship hosted by the TUM School of Social Sciences and Technology and the TUM Think Tank. We sincerely acknowledge the financial support provided by the fellowship. Additionally, we would like to extend our gratitude to the TUM Data Analytics\&Statistics chair, under the leadership of Alexander Fraser.

\bibliography{anthology,custom}

\onecolumn
\appendix

\section{English SOTA models for the Tasks}
\label{sec:app_english_sota}
We used the following publicly available instances of already fine-tuned English models for the considered tasks: (i) toxicity classifier\footnote{\href{https://huggingface.co/martin-ha/toxic-comment-model}{https://huggingface.co/martin-ha/toxic-comment-model}}; (ii) formality classifier~\footnote{\href{https://huggingface.co/cointegrated/roberta-base-formality}{https://huggingface.co/cointegrated/roberta-base-formality}}; (iii) NLI classifier~\footnote{\href{https://huggingface.co/cross-encoder/nli-deberta-base}{https://huggingface.co/cross-encoder/nli-deberta-base}}.

Not for all within these models, the report of train/val/test splits usage was provided. Thus, we cannot fairly test translated into English input with these models, only natural test sets.

\section{LLMs Exploration for Ukrainian Texts Classification}
\label{sec:app_llms_exploration}
In the course of the work, four LLMs were considered: Llama-2 \footnote{\href{https://huggingface.co/meta-llama/Llama-2-7b-chat-hf}{https://huggingface.co/meta-llama/Llama-2-7b-chat-hf}}~\cite{touvron2023llama}, LLaMa-3\footnote{\href{https://huggingface.co/meta-llama/Meta-Llama-3.1-8B-Instruct}{https://huggingface.co/meta-llama/Meta-Llama-3.1-8B-Instruct}}~\cite{llama3modelcard}, Mistral\footnote{\href{https://huggingface.co/mistralai/Mistral-7B-v0.1}{https://huggingface.co/mistralai/Mistral-7B-v0.1}}~\cite{Jiang2023Mistral7} and FLAN-T5\footnote{\href{https://huggingface.co/google/flan-t5-base}{https://huggingface.co/google/flan-t5-base}}~\cite{chung2022scaling}.
The most important task for this section of the research was to find the optimal prompt. For Llama-2,3 and Mistral, the prompts that showed the best results are presented in Appendix~\ref{sec:app_llm_prompts}. For the toxicity classification task, the labels ``toxic'' and ``non-toxic'' were initially used, but later they were changed to ``obscene'' and ''normal'' to improve the results and this contributed to an increase in accuracy. For the NLI and formality classification tasks, the type of problem to be solved was added to the prompt along with examples, and while Mistral for NLI showed good results, for the formality task, due to the fact that the translation of data from English blurs the boundaries between labels, a satisfactory result was not yet achieved.

FLAN-T5, on the other hand, despite being trained on other Slavic languages, did not show the desired result for Ukrainian. Nevertheless, we tested the English prompts for classification tasks, and the model showed a decent result, so it was considered for backtranslation task. It showed compatible results with other bigger models for classification tasks. However, for Ukrainian, the performance was not peak.

\begin{table*}[ht!]
\footnotesize
\centering

\begin{tabular}{p{3.65cm}|c|c|c|c|c|c|c|c}
\toprule

& \textbf{Acc} & \textbf{Pr} & \textbf{Re} & \textbf{F1} & \textbf{Acc} & \textbf{Pr} & \textbf{Re} & \textbf{F1} \\ \hline

& \multicolumn{4}{c|}{{Translated Test Set}} & \multicolumn{4}{c}{{Semi-natural Test Set}} \\ \hline \hline

\multicolumn{9}{c}{\textbf{Toxicity Classification}} \\ \hline

LLaMa-2 Prompting & 0.51 & 0.50 & 0.67 & 0.42 & 0.67 & 0.67 & 0.49 & 0.67  \\
LLaMa-3 Prompting & 0.61 & 0.56 & 0.66 & 0.55 & 0.70 & 0.79 & 0.67 & 0.68  \\
Mistral Prompting &   \textbf{0.86} &
  \textbf{0.68} &
  \textbf{0.74} &
  \textbf{0.70} &
  \textbf{0.76} &
  \textbf{0.81} &
  \textbf{0.76} &
  \textbf{0.75}  \\
FLAN-T5-Backtranslation & \multicolumn{4}{c|}{---}  & 0.69 & 0.73 & 0.69 & 0.68  \\
\hline \hline

\multicolumn{9}{c}{\textbf{Formality Classification}} \\ \hline

LLaMa-2 Prompting & 0.43 & 0.22 & 0.50 & 0.30  & 0.50 & 0.25 & 0.50 & 0.33  \\
LLaMa-3 Prompting & 0.51 & 0.45 & 0.64 & 0.52  & 0.78 & 0.67 & 0.72 & 0.71  \\

Mistral Prompting &     \textbf{0.64} &
  \textbf{0.63} &
  \textbf{0.64} &
  \textbf{0.63} &
  \textbf{0.94} &
  \textbf{0.94} &
  \textbf{0.94} &
  \textbf{0.94}  \\
FLAN-T5-Backtranslation & \multicolumn{4}{c|}{---}  & 0.62 & 0.77 & 0.62 & 0.56  \\

\hline \hline

\multicolumn{9}{c}{\textbf{Natural Language Inference}} \\ \hline

LLaMa-2 Prompting & 0.36 & 0.40 & 0.36 & 0.34  & 0.37 & 0.28 & 0.36 & 0.28  \\
LLaMa-3 Prompting & 0.55 & 0.50 & \textbf{0.57} & 0.55  & 0.66 & 0.68 & 0.66 & 0.66  \\
Mistral Prompting &     \textbf{0.56} &
  \textbf{0.61} &
 0.56 &
  \textbf{0.56} &
  \textbf{0.71} &
  \textbf{0.72} &
  \textbf{0.69} &
  \textbf{0.69}  \\
FLAN-T5-Backtranslation & \multicolumn{4}{c|}{---} & 0.48 & 0.68& 0.49 & 0.42  \\

\bottomrule
\end{tabular}
\caption{Ukrainian Texts Classification results using Large Language Models.} 
\label{tab:llms_results}
\end{table*}

\clearpage
\newpage

\section{LLM Tasks Prompts}
Here, we provide the full listing of the prompts used to obtain the results from LLMs.

\label{sec:app_llm_prompts}

\paragraph{Toxicity Classification} \textit{\newline Classify the text into two categories: contains obscene words or contains none obscene words. Reply with only one word: obscene or normal.
\newline
\newline
Examples:
\newline
Text: \foreignlanguage{ukrainian}{Сьогодні знайти у відкритих джерелах точну суму, витрачену на будівництво об’єкта, що про нього мова, майже неможливо.}
Sentiment: normal.
\newline
Text: \foreignlanguage{ukrainian}{знаєте, якщо свої дебільні коментарі ще й ілюструвати посиланнями на російську вікі, хтось може здогадатися, що ви тупий єблан.}
Sentiment: obscene.
\newline
\newline
Text: \{text\}
\newline
Sentiment:
}

\paragraph{Formality Classification} \textit{\newline This is detecting text formality task. Classify the text into two categories: formal or informal. 
\newline
\newline
Examples:
\newline
Text: \foreignlanguage{ukrainian}{У вас вже є остаточне рішення щодо кольору весільної сукні?}
Sentiment: formal.
\newline
Text: \foreignlanguage{ukrainian}{Незважаючи на те шо було до цього, знаєте що, я думаю, що тобі все ж таки слід зробити перший крок!}
Sentiment: informal.
\newline
\newline
Text: \{text\}
\newline
Sentiment:}

\paragraph{Natural Language Inference} \textit{\newline This is Natural language inference (NLI) task. Determine whether a given hypothesis is contradiction, entailment or neutral in relation to a given premise. Reply with only one word: contradiction, neutral or entailment.
\newline
\newline
Examples:
\newline
Premise: \foreignlanguage{ukrainian}{Чоловік у чорній сорочки грає в гольф ззовні.}
Hypothesis: \foreignlanguage{ukrainian}{Чоловік грає на полі гольфу, щоб відпочити.}
Label: neutral.
\newline
Premise: \foreignlanguage{ukrainian}{Чоловік у чорній сорочки грає в гольф ззовні.}
Hypothesis: \foreignlanguage{ukrainian}{Чоловік у чорній сорочки обмінюється картами з дівчиною.}
Label: contradiction.
\newline
Premise: \foreignlanguage{ukrainian}{Чоловік у чорній сорочки грає в гольф ззовні.}
Hypothesis: \foreignlanguage{ukrainian}{Чоловік у чорній сорочки грає в гольф.}
Label: entailment
\newline
\newline
Text: \{premise, hypothesis\}
\newline
Label:}

\clearpage
\newpage

\section{Translated Data Examples}
\label{sec:app_translated_data}

Here, we provide examples of Ukrainian texts obtained by translation from English data for the corresponding tasks: (i) toxicity classification in Table~\ref{tab:app_toxicity_translated_examples}; (ii) formality classification in Table~\ref{tab:app_formality_translated_examples}; (iii) NLI classification in Table~\ref{tab:app_nli_translated_examples}. For English texts, the original grammar, as presented in the source data, is retained.

\begin{table}[h!]
    \centering
    \footnotesize
    \begin{tabular}{p{2cm}|p{11cm}}
    \toprule
        
        Toxic & \foreignlanguage{ukrainian}{ви всі тупі осли.} \newline \textcolor{gray}{\scriptsize{\textit{youre all dumb asses}}} \\ 
        \midrule 
        Non-toxic & \foreignlanguage{ukrainian}{Є два адміністратори, які досить добре працюють з такими статтями, можливо, ви могли б зв'язатися з ними.} \newline \textcolor{gray}{\scriptsize{\textit{there are two admins that do handle such articles pretty well you could maybe contact  and }}} \\
        \midrule 
        Toxic & \foreignlanguage{ukrainian}{І Роберт - це чорне лайно} \newline \textcolor{gray}{\scriptsize{\textit{and robert is a black shit}}} \\ 
        \midrule 
        Non-toxic & \foreignlanguage{ukrainian}{Гаразд, я почав трансляцію нової статті, я використав вашу запропоновану назву може використовувати більше деталей зараз} \newline \textcolor{gray}{\scriptsize{\textit{ok i started the transtion the new article i used your suggested title could use a lot more detail now}}} \\
        \midrule 
        Toxic & \foreignlanguage{ukrainian}{Що за купа ср*них ботанів?} \newline \textcolor{gray}{\scriptsize{\textit{what a bunch of f**king nerds}}} \\ 
        \midrule 
        Non-toxic & \foreignlanguage{ukrainian}{Зупиніться, будь ласка, якщо ви продовжите вандализувати сторінки, ви будете заблоковані від редагування wikipedia} \newline \textcolor{gray}{\scriptsize{\textit{please stop if you continue to vandalize wikipedia you will be blocked from editing}}} \\
        \midrule 
        Toxic & \foreignlanguage{ukrainian}{Альтернативна поп-культура, що означає п*зда чи ци, розкидає таємничу сучку, яка руйнує все, що примара називає когось, це спосіб дати людині знати, що вони є п*зда в той час як цензують інших навколо вас в громадських місцях або в соціальних кутах, сучасний сленг попереджаючи інших про небезпеку.} \newline \textcolor{gray}{\scriptsize{\textit{alternative  pop culture meaning   c*nt or cee unt  a percieved mysterious bitch that destroys everything  whem calling someone this is a way of letting anyone know they are a c*nt while censoring others around you in public or in social corners  a modern slang alerting other of the danger}}} \\ 
        \midrule 
        Non-toxic & \foreignlanguage{ukrainian}{Адміністратори виконують дії, що ґрунтуються на громадському консенсусі, вони не приймають односторонніх рішень далі, тому у зв'язку з цим редактори, які зосереджують свою увагу на виборах або канадалях, не мають можливості перенаправити кандидатів на партійні статті.} \newline \textcolor{gray}{\scriptsize{\textit{  admins execute actions based on community consensus  they do not make unilateral decisions further that afd did not have the involvement of editors who focus on ontario or canadawide elections so they were likely unfamiliar with the option of redirecting to party candidate articles}}} \\
        \bottomrule
    \end{tabular}
    \caption{Examples of the translated samples for the \textbf{Toxicity Classification} task. English original sentences are taken from the Jigsaw dataset~\cite{jigsaw}.}
    \label{tab:app_toxicity_translated_examples}
\end{table}

\begin{table}[h!]
    \centering
    \footnotesize
    \begin{tabular}{p{2cm}|p{11cm}}
    \toprule
        
        Formal & \foreignlanguage{ukrainian}{Тільки тому, що він має потенціал бути гідним хлопцем, це не означає, що він стане.} \newline \textcolor{gray}{\scriptsize{\textit{Just because he has potential to be a decent boyfriend, does not mean that he will be.}}} \\
        \midrule
        Informal & \foreignlanguage{ukrainian}{Вам удачі!} \newline \textcolor{gray}{\scriptsize{\textit{The Best of Luck to ya!}}} \\
        \midrule
        Formal & \foreignlanguage{ukrainian}{Будь-яка жінка виглядає привабливо, коли стоїть поруч з непривабливим чоловіком.} \newline \textcolor{gray}{\scriptsize{\textit{Any woman looks attractive when standing beside an unattractive man.}}} \\
        \midrule
        Informal & \foreignlanguage{ukrainian}{Це найглупше, що я чув.} \newline \textcolor{gray}{\scriptsize{\textit{thats the stupidest thing I have ever heard.}}} \\
        \midrule
        Formal & \foreignlanguage{ukrainian}{О, мій улюблений класичний телешоу - "Золоті дівчата". Чи цей шоу вважається "класичним"?} \newline \textcolor{gray}{\scriptsize{\textit{Oh, my favorite classic television show is 'The Golden Girls.' Is that show considered a 'classic'?}}} \\
        \midrule
        Informal & \foreignlanguage{ukrainian}{Я б, ймовірно, випив все в проклятому барі, щоб навіть подумати про те, що знайду тебе привабливим.} \newline \textcolor{gray}{\scriptsize{\textit{i'd probably drink up everything in the damn bar to even think of finding u attractive.}}} \\
        \midrule
        Formal & \foreignlanguage{ukrainian}{Я також володію компакт-диском Vanilla Ice.} \newline \textcolor{gray}{\scriptsize{\textit{I also own the Vanilla Ice compact disk.}}} \\
        \midrule
        Informal & \foreignlanguage{ukrainian}{LOL просто граю, але вони супер h-o-t} \newline \textcolor{gray}{\scriptsize{\textit{lol just playin but they are super h-o-t}}} \\
    
        \bottomrule
    \end{tabular}
    \caption{Examples of the translated samples for the \textbf{Formality Classification} task. English original sentences are taken from the GYAFC dataset~\cite{rao-tetreault-2018-dear}.}
    \label{tab:app_formality_translated_examples}
\end{table}

\begin{table}[h!]
    \centering
    \footnotesize
    \begin{tabular}{p{2cm}|p{11cm}}
    \toprule
        
        Premise \newline \newline Hypothesis \newline \newline Label & \foreignlanguage{ukrainian}{Маленький хлопчик ходить по трубі, яка протягується над водою.} \newline \textcolor{gray}{\scriptsize{\textit{A young boy walks on a pipe that stretches over water.}}} \newline\foreignlanguage{ukrainian}{Хлопчик ризикує впасти в воду.} \newline \textcolor{gray}{\scriptsize{\textit{A boy is in danger of falling into water.}}} \newline neutral\\
        \midrule 
        Premise \newline \newline Hypothesis \newline \newline Label & \foreignlanguage{ukrainian}{Деякі собаки бігують на пустельному пляжі.} \newline \textcolor{gray}{\scriptsize{\textit{Some dogs are running on a deserted beach.}}} \newline\foreignlanguage{ukrainian}{Вони на Гавайях.} \newline \textcolor{gray}{\scriptsize{\textit{They are in Hawaii.}}} \newline neutral\\
        \midrule 
        Premise \newline \newline Hypothesis \newline \newline Label & \foreignlanguage{ukrainian}{Два чоловіки стоять на човні.} \newline \textcolor{gray}{\scriptsize{\textit{Two men are standing in a boat.}}} \newline\foreignlanguage{ukrainian}{Деякі чоловіки стоять на вершині машини.} \newline \textcolor{gray}{\scriptsize{\textit{Some men are standing on top of a car.}}} \newline contradiction\\
        \midrule 
        Premise \newline \newline Hypothesis \newline \newline Label & \foreignlanguage{ukrainian}{Мотоциклетні гонки.} \newline \textcolor{gray}{\scriptsize{\textit{A biker races.}}} \newline\foreignlanguage{ukrainian}{Автомобіль жовтий.} \newline \textcolor{gray}{\scriptsize{\textit{The car is yellow}}} \newline contradiction\\
        \midrule 
        Premise \newline \newline Hypothesis \newline \newline Label & \foreignlanguage{ukrainian}{Чоловік у синій куртці вирішив лежати на траві.} \newline \textcolor{gray}{\scriptsize{\textit{Male in a blue jacket decides to lay in the grass.}}} \newline\foreignlanguage{ukrainian}{Чоловік у синій куртці лежить на зеленої траві.} \newline \textcolor{gray}{\scriptsize{\textit{The guy wearing a blue jacket is laying on the green grass}}} \newline entailment\\
        \midrule 
        Premise \newline \newline Hypothesis \newline \newline Label & \foreignlanguage{ukrainian}{Людина, яка сидить на скелі біля водопаду.} \newline \textcolor{gray}{\scriptsize{\textit{A person sitting on a rock beside a waterfall.}}} \newline\foreignlanguage{ukrainian}{Людина знаходиться біля води.} \newline \textcolor{gray}{\scriptsize{\textit{A person is near water}}} \newline entailment\\
        \bottomrule
    \end{tabular}
    \caption{Examples of the translated samples for the \textbf{Natural Language Inference} task. English original sentences are taken from the SNLI dataset~\cite{snliemnlp2015}.}
    \label{tab:app_nli_translated_examples}
\end{table}

\clearpage
\newpage

\section{Semi-natural Test Data Examples}
\label{sec:app_natural_data}

Here, we provide examples from the natural Ukrainian texts obtained for the corresponding tasks: (i) toxicity classification in Table~\ref{tab:app_toxicity_natural_examples}; (ii) formality classification in Table~\ref{tab:app_formality_natural_examples}; (iii) NLI classification in Table~\ref{tab:app_nli_natural_examples}.

\begin{table}[h!]
    \centering
    \footnotesize
    \begin{tabular}{p{2cm}|p{11cm}}
    \toprule
        
        Toxic & \foreignlanguage{ukrainian}{@pfactum нє, китай рулить, то однозначно. ден сяопін був генієм економіки. але це було підписано бо більше ні на шо пі**рович не заслужив:)} \newline \textcolor{gray}{\scriptsize{\textit{@pfactum no, the Chinese drive, of course. The shoopin was an economic genius. But it was signed because no more on the sublarcier was worthy of:)}}} \\ 
        \midrule 
        Non-toxic & \foreignlanguage{ukrainian}{@G1NTONIK 1) доброго часу, коліжанці дав почитати збірку, багато що оцінила, але запитала про "Самонедостатність" ..} \newline \textcolor{gray}{\scriptsize{\textit{@G1NTONIK 1) good time, the colts gave the collection a lot of reading and appreciated, but asked about "Memonysity..."}}} \\
        \midrule 
        Toxic & \foreignlanguage{ukrainian}{вже не пі**рас?} \newline \textcolor{gray}{\scriptsize{\textit{Isn't that a f**got?}}} \\ 
        \midrule 
        Non-toxic & \foreignlanguage{ukrainian}{Не раз заявляв про наміри зайти на наш ринок ірландський Ryanair .} \newline \textcolor{gray}{\scriptsize{\textit{More than once, he claimed to visit our market in Irish Ryanair.}}} \\
        \midrule 
        Toxic & \foreignlanguage{ukrainian}{сьогоднішня мрія - адекватний транспорт в крим, щоб не доводилося щоразу мозок собі ї**ти стиковкою цих жахливих людиноненависницьких рейсів} \newline \textcolor{gray}{\scriptsize{\textit{Today's dream is a safe transport into the ice so that every brain doesn't have to f**k its way through these terrible man - hated flights.}}} \\ 
        \midrule 
        Non-toxic & \foreignlanguage{ukrainian}{Співрозмовники досягли домовленості про проведення чергового засідання Спільної міжурядової українсько - туркменської комісії з економічного та культурно - гуманітарного співробітництва вже ближчим часом .} \newline \textcolor{gray}{\scriptsize{\textit{Coordinators have reached an agreement to hold a joint Intergovernmental Union Commission on Economic and Cultural Cooperation for a longer time.}}} \\
        \midrule 
        Toxic & \foreignlanguage{ukrainian}{нема відчуття гіршого, ніж коли розумієш, шо ти конкретно так тупанув, і через це все йде по п**ді.} \newline \textcolor{gray}{\scriptsize{\textit{There's no worse feeling than when you realize that you were exactly f**king that way, and that's why everything goes on p*ss.}}} \\ 
        \midrule 
        Non-toxic & \foreignlanguage{ukrainian}{Державне підприємство « Конструкторське бюро „ Південне “ ім . М . К . Янгеля » було створено 1951 як конструкторський відділ Південного машинобудівного заводу з виробництва військових ракет .} \newline \textcolor{gray}{\scriptsize{\textit{The state enterprise (C) was created by 1951 as the South Carworker's design department for the production of military rockets.}}} \\
        \bottomrule
    \end{tabular}
    \caption{Examples of the natural samples for the \textbf{Toxicity Classification} task obtained from Ukrainian tweets corpus from~\cite{bobrovnik2019twt} and news and fiction UD Ukrainian IU dataset~\cite{udiu2016}.}
    \label{tab:app_toxicity_natural_examples}
\end{table}

\begin{table}[h!]
    \centering
    \footnotesize
    \begin{tabular}{p{2cm}|p{11cm}}
    \toprule
        
        Formal & \foreignlanguage{ukrainian}{Повноваження судді Конституційного Суду та гарантії його діяльності не можуть бути обмежені при введенні воєнного чи надзвичайного стану в Україні або в окремих її місцевостях.} \newline \textcolor{gray}{\scriptsize{\textit{The powers of the Constitutional Court Judge and the guarantees of his activity may not be limited when martial law or emergency is imposed in Ukraine or in certain areas of Ukraine.}}} \\
        \midrule
        Informal & \foreignlanguage{ukrainian}{я навіть не знаю, хто біля них зупиняється О\_о якийсь зоопарк між Чернігівською і Лісовою. траса. ну ви поняли ;)} \newline \textcolor{gray}{\scriptsize{\textit{I don't even know who's staying near them. Some kind of zoo between the Chernigov and the Forest.}}} \\
        \midrule
        Formal & \foreignlanguage{ukrainian}{Суд упродовж трьох місяців з дня офіційного опублікування цього Закону ухвалює Регламент та утворює сенати у порядку, встановленому цим Законом.} \newline \textcolor{gray}{\scriptsize{\textit{The Court shall, within three months of the date of the official publication of this Act, adopt a regulation and set up the Senate in the manner provided for by this Act.}}} \\
        \midrule
        Informal & \foreignlanguage{ukrainian}{так за сьогодні находилась, шо мене аж тошнить від втоми. чуствую, шо коли доповзу в ліжку просто вмру на місці} \newline \textcolor{gray}{\scriptsize{\textit{I'm so tired, I feel like I'm gonna die in bed.}}} \\
        \midrule
        Formal & \foreignlanguage{ukrainian}{Коли це доцільно, держави-члени співпрацюють з метою об'єднання організаційних зусиль для спільних дій.} \newline \textcolor{gray}{\scriptsize{\textit{Member States shall cooperate to pool their organisational efforts for joint action where appropriate.}}} \\
        \midrule
        Informal & \foreignlanguage{ukrainian}{я значить сиджу така серйозна, перевіряю зошити, а тут це)) дякую, я навіть спати перехотіла від сміху))} \newline \textcolor{gray}{\scriptsize{\textit{I mean, I'm sitting so serious, checking to see if I'm getting any sleep, and here it is.}}} \\
        \bottomrule
    \end{tabular}
    \caption{Examples of the natural samples for the \textbf{Formality Classification} task obtained from Ukrainian legal acts~\cite{legaltestuk2012} and EU acts in Ukrainian~\cite{euactsukr2012}.}
    \label{tab:app_formality_natural_examples}
\end{table}

\begin{table}[h!]
    \centering
    \footnotesize
    \begin{tabular}{p{2cm}|p{11cm}}
    \toprule
        
        Premise \newline \newline Hypothesis \newline \newline Label & \foreignlanguage{ukrainian}{Архів Суду} \newline \textcolor{gray}{\scriptsize{\textit{Court Archives}}} \newline\foreignlanguage{ukrainian}{Матеріали діяльності Суду зберігаються в Архіві Суду.} \newline \textcolor{gray}{\scriptsize{\textit{The Court's activities are kept in the Court's Archives.}}} \newline entailment\\
        \midrule 
        Premise \newline \newline Hypothesis \newline \newline Label & \foreignlanguage{ukrainian}{Між тим актори місяцями не одержували грошового утримання} \newline \textcolor{gray}{\scriptsize{\textit{Meanwhile, the actors have been without a cash stipend for months}}} \newline\foreignlanguage{ukrainian}{Перше всього загомоніли кого тепер у предводителі} \newline \textcolor{gray}{\scriptsize{\textit{First, they've gotten who's in charge.}}} \newline neutral\\
        \midrule 
        Premise \newline \newline Hypothesis \newline \newline Label & \foreignlanguage{ukrainian}{Мети походу досягнуто тож не гаючись назад} \newline \textcolor{gray}{\scriptsize{\textit{The goal of the campaign was achieved without turning back}}} \newline\foreignlanguage{ukrainian}{В деяких місцях вітер позносив з піль сніг в інших знову лежали хвилясті замети} \newline \textcolor{gray}{\scriptsize{\textit{In some places the winds carried away the snow from the mud in others again lay wavy notes}}} \newline neutral\\
        \midrule 
        Premise \newline \newline Hypothesis \newline \newline Label & \foreignlanguage{ukrainian}{Вода замерзає при нагріванні} \newline \textcolor{gray}{\scriptsize{\textit{Water freezes when heated}}} \newline\foreignlanguage{ukrainian}{Вода кипить при нагріванні} \newline \textcolor{gray}{\scriptsize{\textit{Water boils when heated}}} \newline contradiction\\
        \midrule 
        Premise \newline \newline Hypothesis \newline \newline Label & \foreignlanguage{ukrainian}{Психологічна стійкість - це ключ до подолання будь-яких перешкод} \newline \textcolor{gray}{\scriptsize{\textit{Psychological resilience is the key to overcoming any obstacle}}} \newline\foreignlanguage{ukrainian}{Психологічна стійкість може бути причиною втрати емоційної чутливості та співчутливості} \newline \textcolor{gray}{\scriptsize{\textit{Psychological resilience can cause loss of emotional sensitivity and empathy}}} \newline contradiction\\
        \bottomrule
    \end{tabular}
    \caption{Examples of the natural samples for the \textbf{Natural Language Inference} task obtained from Ukrainian legal acts~\cite{legaltestuk2012}, EU acts in Ukrainian~\cite{euactsukr2012}, UberText 2.0~\cite{chaplynskyi-2023-introducing}, and contradiction label samples were constructed by the Ukrainian native speaker.}
    \label{tab:app_nli_natural_examples}
\end{table}

\end{document}